%% file: main.tex
\documentclass[10pt,twocolumn,letterpaper]{article}

\usepackage[pagenumbers]{cvpr} 


\definecolor{cvprblue}{rgb}{0.21,0.49,0.74}
\usepackage[pagebackref,breaklinks,colorlinks,allcolors=cvprblue]{hyperref}
\usepackage{amsmath,amssymb}
\usepackage{algorithm} 
\usepackage{algpseudocode} 
\usepackage{booktabs} 
\usepackage{array} 
\usepackage{float}
\usepackage{graphicx} 
\usepackage{caption}
\usepackage{multicol} 
\usepackage[utf8]{inputenc}  
\usepackage[T1]{fontenc}     
\usepackage{CJKutf8}        
\usepackage{multirow}
\usepackage{multicol}
\usepackage{makecell}
\usepackage{bbding}

\def\confName{CVPR}
\def\confYear{2026}

\title{FaithFusion: Harmonizing Reconstruction and Generation via Pixel-wise Information Gain}

\author{
\textbf{YuAn Wang}$^{1^{*}}$\quad
\textbf{Xiaofan Li}$^{1^{*,\dagger}}$\quad
\textbf{Chi Huang}$^{1}$\quad
\textbf{Wenhao Zhang}$^{1,2}$\\
\textbf{Hao Li}$^{1}$\quad
\textbf{Bosheng Wang}$^{1}$\quad
\textbf{Xun Sun}$^{1}$\quad
\textbf{Jun Wang}$^{1}$\\[2pt]
$^{1}$Baidu Inc.\quad
$^{2}$Nanjing University\\[2pt]
\\
{Project page: \url{https://shalfun.github.io/faithfusion}}
\\ 
}

\begin{document}

\maketitle
\renewcommand{\thefootnote}{\fnsymbol{footnote}}
\footnotetext[1]{Equal contribution.}
\footnotetext[2]{Corresponding author: \texttt{\small Shalfunnn@gmail.com}}
\begin{abstract}
In controllable driving-scene reconstruction and 3D scene generation, maintaining geometric fidelity while synthesizing visually plausible appearance under large viewpoint shifts is crucial. However, effective fusion of geometry-based 3DGS and appearance-driven diffusion models faces inherent challenges, as the absence of pixel-wise, 3D-consistent editing criteria often leads to over-restoration and geometric drift. To address these issues, we introduce \textbf{FaithFusion}, a 3DGS-diffusion fusion framework driven by pixel-wise Expected Information Gain (EIG). EIG acts as a unified policy for coherent spatio-temporal synthesis: it guides diffusion as a spatial prior to refine high-uncertainty regions, while its pixel-level weighting distills the edits back into 3DGS. The resulting plug-and-play system is free from extra prior conditions and structural modifications.Extensive experiments on the Waymo dataset demonstrate that our approach attains SOTA performance across NTA-IoU, NTL-IoU, and FID, maintaining an FID of 107.47 even at 6 meters lane shift. Our code is available at https://github.com/wangyuanbiubiubiu/FaithFusion.
\end{abstract}

\begin{figure}[!htbp]
    \centering
    \includegraphics[width=1.0\linewidth]{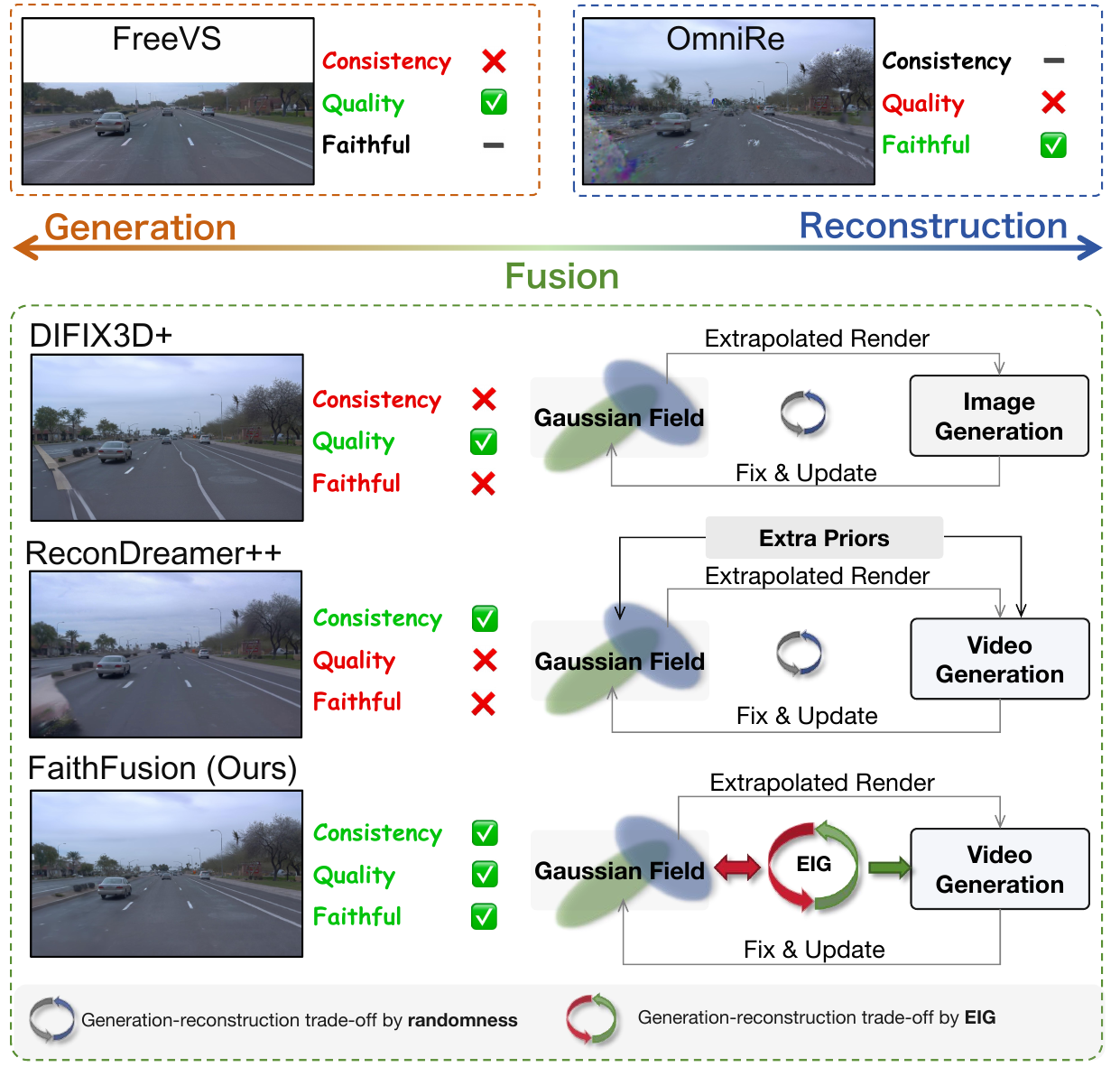}
    \caption{\textbf{Comparative overview.} 
    Comparison of FreeVS~\cite{wang2024freevs}, OmniRe~\cite{chen2024omnire}, the fusion-based methods DIFIX3D+~\cite{wu2025difix3d+} and ReconDreamer++~\cite{zhao2025recondreamer++}, and our EIG-integrated \emph{FaithFusion}, which simultaneously achieves consistency, quality, and faithfulness.
    }
    \label{fig:main}
\end{figure}
\vspace{-12pt}
\section{Introduction}
\label{sec:intro}
Building a controllable driving world for closed-loop simulation~\cite{tonderski2024neurad,li2024drivingdiffusion,yang2023unisim} requires jointly achieving geometric fidelity in reconstruction and controllability in appearance generation. Neural rendering, particularly Neural Radiance Fields (NeRF)~\cite{mildenhall2021nerf,barron2021mip,guo2023streetsurf} and 3D Gaussian Splatting (3DGS)~\cite{kerbl20233d,lu2024scaffold,kheradmand20243d}, has transformed 3D representations and novel view synthesis. However, under sparse observations, heavy occlusions, or viewpoints far from the training trajectory, NeRF and 3DGS often yield geometric inconsistencies and artifacts~\cite{zhang2024cor,chen2024pgsr,yang2025gurecon}. Diffusion models~\cite{ho2020denoising,song2023consistency,lipman2022flow} excel at image and video generation and restoration, yet without pixel-level, geometry-consistent guidance, they tend to cause over-restoration and introduce geometric drift. Thus, balancing faithful reconstruction with controllable generation remains a central challenge for unified world modeling~\cite{yu2024sgd,zhao2025drivedreamer4d,zhao2025recondreamer++}.

Unlike existing reconstruction or generation methods,
fusion reconstruction–generation approaches primarily adopt an online progressive loop paradigm, following a ``render–restoration–feedback'' workflow designed to minimize information loss during 3DGS rendering, as shown in \cref{fig:main}. Building upon this foundation, current efforts focus on optimizing both the generation and reconstruction branches: the former aims to improve temporal consistency via extra prior conditions~\cite{wang2024freevs,ni2025recondreamer}, while DIFIX3D+~\cite{wu2025difix3d+} achieves a balance between efficiency and quality through single-step, single-frame restoration. The reconstruction branch involves structural modifications to the 3DGS architecture or the introduction of robust geometry priors~\cite{zhao2025recondreamer++}. However, these explorations face common limitations: they predominantly rely on view-level heuristics to decide ``where, when, and how much to edit''. This dependence on coarse-grained guidance directly leads to insufficient control over generation; once diffusion is activated, it often overwrites already correct regions, causing over-restoration and geometric drift. Therefore, a principled decision mechanism that determines which regions to generate and which to preserve is still absent.

\begin{figure*}[htbp]
    \centering
    \includegraphics[width=\linewidth]{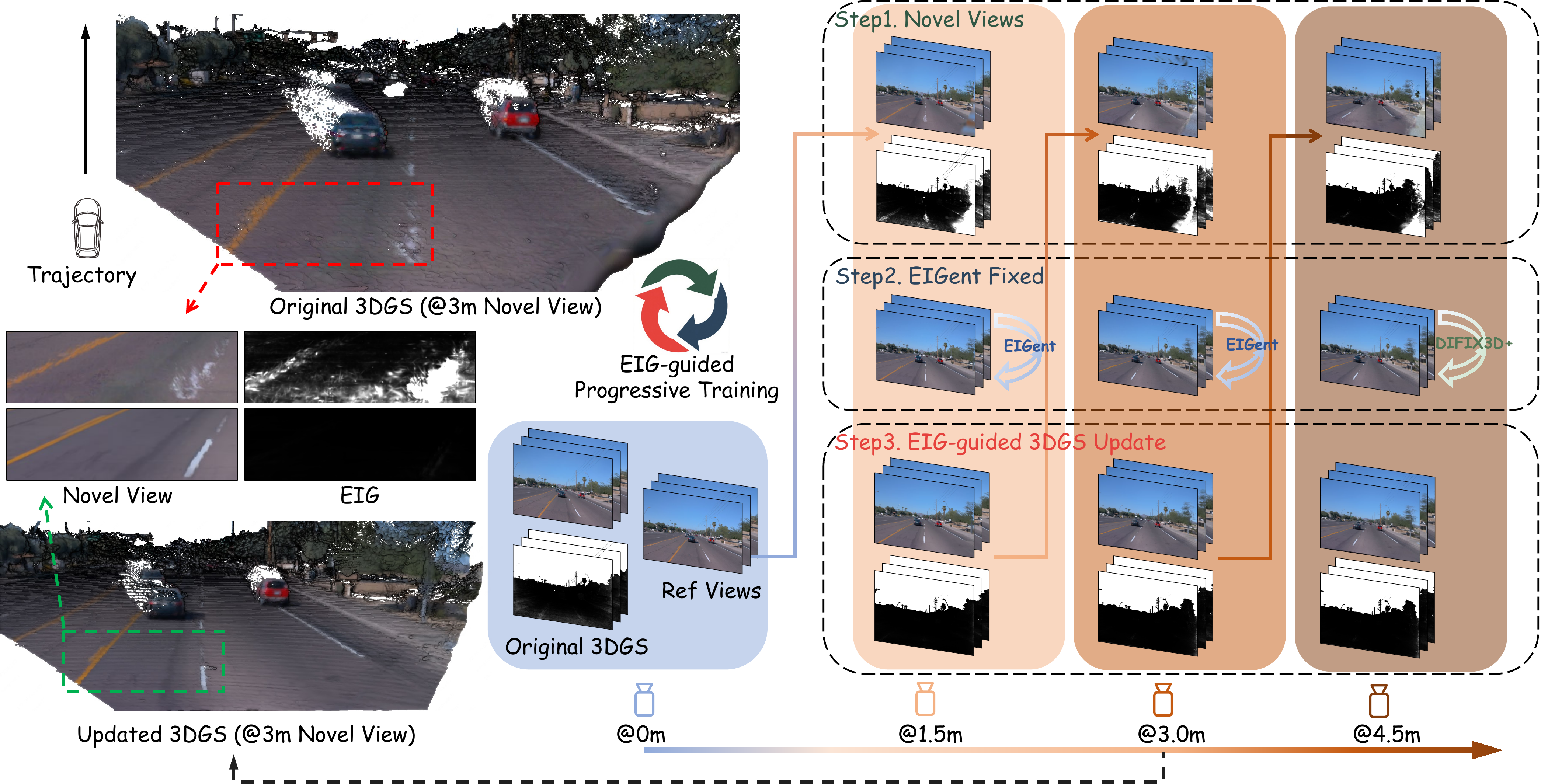}
    \caption{\textbf{FaithFusion pipeline.}
    The EIG-guided progressive training loop with three steps:
    \textbf{Step 1:} \emph{Novel-view synthesis}. Render laterally offset novel views and their pixel-level EIG maps from the original 3DGS.
    \textbf{Step 2:} \emph{EIGent Fixed}. Feed the renders and EIG maps into EIGent to repair high-EIG regions---using Video DiT early for spatio-temporal consistency and DIFIX3D+ later for per-frame perceptual refinement.
    \textbf{Step 3:} \emph{EIG-guided 3DGS Update.} Fine-tune the 3DGS model with the EIGent-restored views and EIG maps.
    }
    \label{fig:faithfusion-pipline}
\end{figure*}

Motivated by these limitations, we introduce \textbf{FaithFusion}, a 3DGS-diffusion fusion paradigm driven by pixel-wise Expected Information Gain (EIG). 
Our core insight is to reformulate the decision of whether and how much to edit a pixel into a forward-looking information-theoretic metric---how much the edit reduces posterior uncertainty. 
To this end, we tightly couple a Laplace-approximated EIG with the differentiable 3DGS renderer to derive a pixel-level estimation specifically tailored for it. 
As illustrated in \cref{fig:faithfusion-pipline}, \emph{FaithFusion} builds upon this foundation by:

\begin{itemize}
   \item \textbf{Generation branch:} 
    EIG serves as a spatial weighting function that guides diffusion to generate content only in high-uncertainty regions, effectively suppressing over-restoration and geometric drift.
    \item \textbf{Reconstruction branch:} 
    EIG functions as a pixel-wise loss weight, progressively distilling high-value edits back into 3DGS and forming a unified loop of edit triggering, strength modulation, and knowledge feedback.
    \item \textbf{System properties:} 
    Without relying on external priors or modifying the 3DGS architecture, \emph{FaithFusion} preserves trajectory fidelity while significantly improving spatio-temporal consistency and perceptual quality under large viewpoint shifts such as lane changes.
\end{itemize}

We conduct comprehensive evaluations on the Waymo dataset~\cite{sun2020scalability}, achieving substantial improvements on NTA-IoU, NTL-IoU, and FID metrics. 
Extensive visualizations demonstrate precise corrections in under-constrained regions, minimal disturbance to original trajectories, and consistent behavior across diverse viewpoints. 
In summary, \emph{FaithFusion} replaces heuristic decisions about ``where, when, and how much to edit'' with a principled, information-theoretic formulation: offering a concise, interpretable, and generalizable framework for unified, controllable 3D scene modeling.

\vspace{-5pt}
\section{Related Work}
\label{sec:related}
\textbf{Driving Scene Reconstruction.}
The introduction of NeRF~\cite{mildenhall2021nerf,muller2022instant,barron2023zip} and  increasingly influential 3DGS~\cite{kerbl20233d,yu2024mip,lu2024scaffold} has brought high-quality novel view synthesis (NVS) to the forefront of autonomous driving research. A primary challenge is coping with the ubiquitous dynamics in driving scenes. Approaches differ in how they handle dynamic content: self-supervised or weakly supervised methods, whether NeRF-based~\cite{yang2023emernerf,turki2023suds} or 3DGS-based~\cite{chen2023periodic,yang2024deformable,huang2024s3gs}, aim to avoid annotation but struggle to robustly disentangle dynamics under motion blur, illumination changes, and other complex cues. 3DGS, thanks to its explicit primitives, naturally supports foreground-background decoupling, and annotation-driven methods~\cite{zhou2024drivinggaussian,yan2024street,zhou2024hugsim,chen2024omnire} are widely adopted. Beyond dynamics, enhancing novel view rendering quality from sparse views remains a key challenge, often tackled through two main strategies. First, researchers introduce geometric priors like LiDAR depth~\cite{sun2024lidarf,zhao2024tclc}, pretrained depth maps~\cite{wang2023sparsenerf,chung2024depth,zhang2024rade,yu2022monosdf}, and driving-specific ground parameterizations~\cite{mei2024rome,guo2023streetsurf,zhou2024hugsim,zhao2025recondreamer++,shen2025og}. However, these priors can inject noise and bias. Second, methods use virtual view augmentation~\cite{cheng2023uc,chen2024pgsr} to enrich observations, which risks local minima due to dependence on accurate warping. Despite progress, large viewpoint shifts (such as lane-change) remain particularly challenging, as shape-radiance ambiguity and unobserved regions still lead to geometric inconsistencies and artifacts.\\
\textbf{Information Gain in Radiance Fields.}
EIG quantifies the value of new observations based on model uncertainty, measuring the uncertainty reduction via information acquisition~\cite{kirsch2022unifying}. Radiance-field uncertainty estimation typically falls into three categories: \emph{variational inference}~\cite{xie2023s,savant2024modeling,pan2022activenerf,hoffman2023probnerf}, which learns probabilistic distributions but demands architectural changes and high training cost; \emph{Monte Carlo sampling}~\cite{savant2024modeling,lyu2024manifold}, which captures uncertainty via sample dispersion but often relies on specific low-dimensional assumptions; and the \emph{Laplace approximation}~\cite{goli2024bayes,jiang2024fisherrf}, a post-hoc, architecture-agnostic, efficient option we adopt for its plug-and-play nature. In novel-view optimization and active mapping, ActiveNeRF~\cite{pan2022activenerf} and FisherRF~\cite{jiang2024fisherrf} already leverage EIG or its heuristics~\cite{liu20243dgs} for view selection, maximizing knowledge gain by measuring uncertainty reduction. Crucially, prior work largely remains at the view level; we extend FisherRF's theory to the pixel level, enabling fine-grained, interpretable diffusion guidance.\\
\textbf{Driving Scenes with Diffusion Priors.} Diffusion models have recently revolutionized image~\cite{ho2020denoising,lipman2022flow,peebles2023scalable} and video~\cite{ho2022video,blattmann2023align,yang2024cogvideox} generation, offering a new paradigm for handling large viewpoint changes in autonomous driving. Coupling 3DGS reconstruction priors with diffusion is often preferred over direct pose-conditioned generation~\cite{zhao2025drivedreamer4d,li2025driverse}, as it reduces the task to restoring degraded novel-view renderings~\cite{wang2024freevs,hwang2024vegs,yu2024sgd,han2025ggs}, which is more tractable. However, these approaches often depend on extra prior conditions, such as LiDAR~\cite{wang2024freevs,yan2025streetcrafter}, sparse annotations~\cite{ni2025recondreamer,chen2025geodrive}, or require 3DGS-specific adaptations to shrink the gap between generated results and reconstruction~\cite{zhao2025recondreamer++}. This dependence limits their generality. In contrast, we propose a novel reconstruction–generation fusion paradigm driven by the intrinsic EIG of 3DGS.  This information-theoretic metric replaces heuristics and strong geometric conditions, significantly enhancing spatio-temporal consistency and perceptual quality under large viewpoint shifts.
\section{Method}
\label{sec:method}
Our primary goal is to develop a method capable of synthesizing high-fidelity, temporally and spatially consistent 4D representations for challenging, far-field novel viewpoints (e.g., lane changes), where traditional 3DGS-based methods often fail. To this end, we propose \emph{FaithFusion}, a controllable 3DGS–diffusion fusion framework driven by pixel-wise Expected Information Gain (EIG). Under a progressive fusion scheme~\cite{haque2023instruct,ni2025recondreamer,wu2025difix3d+}, EIG serves as a unified spatial policy: it first directs the diffusion model to synthesize high-information regions in novel views (e.g., unseen areas), and then guides selective 3DGS fine-tuning to assimilate the generated content, yielding coherent generation–reconstruction coupling.

We describe pixel-wise EIG computation (\cref{sec:3.1}), the dual-branch \emph{EIGent} guidance (\cref{sec:3.2}), and the EIG-driven progressive knowledge integration (\cref{sec:3.3}); the overall pipeline is shown in \cref{fig:faithfusion-pipline}. The framework is \emph{plug-and-play} and can be seamlessly integrated into mainstream street-scene 3DGS systems~\cite{yan2024street,chen2023periodic,yang2024deformable,chen2024omnire}.
\subsection{Expected Information Gain in 3DGS}
\label{sec:3.1}
\textbf{3D Gaussian Splatting Parameterization.} Original 3DGS~\cite{kerbl20233d} represents a static scene with a set of anisotropic Gaussians parameterized by world-space position $\boldsymbol{\mu}_w \in \mathbb{R}^3$, rotation $\mathbf{q}_w \in \mathbb{R}^4$, and scale $\mathbf{s} \in \mathbb{R}^3$. For dynamic street scenes, existing approaches further augment object parameterizations~\cite{yan2024street, chen2023periodic,chen2024omnire,yang2024deformable} to capture motion and deformation. To model view-dependent appearance, each Gaussian maintains spherical harmonic coefficients $\mathbf{c} \in \mathbb{R}^k$ and opacity $o \in \mathbb{R}$. Together, these parameters $\mathbf{\omega}$ yield photorealistic renderings at target timestamps after $\alpha$-blending and projection onto the image plane:
\vspace{-10pt}
\begin{equation}
    \mathbf{C}=\sum_{i \in \mathcal{M}} \mathbf{c}_{i} \alpha_{i}^{\prime} \prod_{j=1}^{i-1}\left(1-\alpha_{j}^{\prime}\right),
    \label{3dgs-alpha-blending}
\end{equation}
where $\mathcal{M}$ denotes the set of Gaussians intersected by the ray, ordered by depth, and $\alpha_i^{\prime}$ is determined by the opacity $o$ and the 2D Gaussian after linear projection.
\begin{figure}[htbp]
    \centering
    \includegraphics[width=0.9\linewidth]{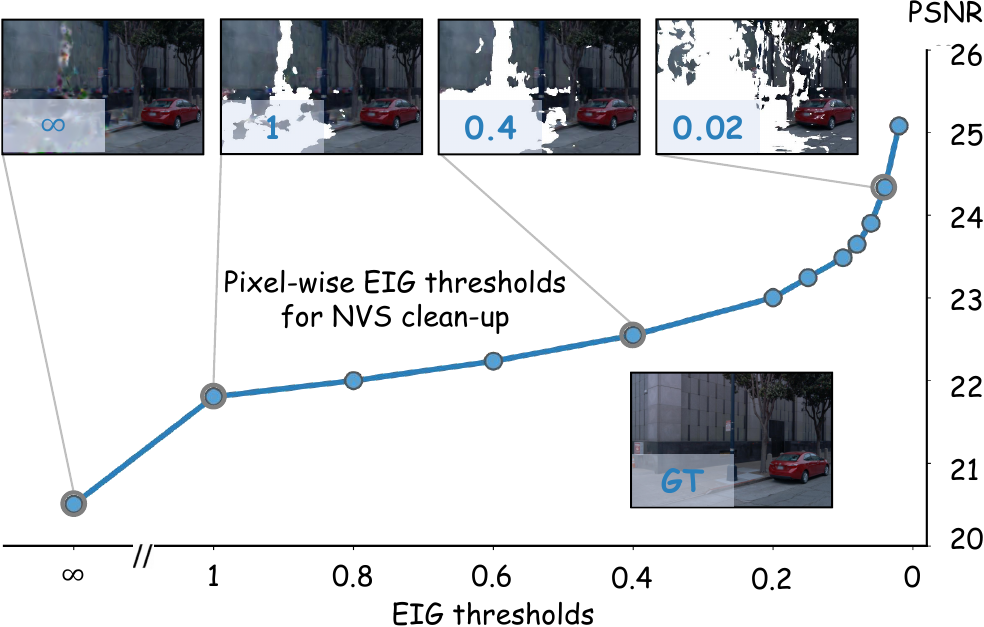}
    \caption{\textbf{Image quality vs. EIG mask threshold.}
    We validate pixel-level EIG as a proxy for novel-view synthesis quality by progressively retaining high-EIG regions and evaluating $\text{PSNR}$.  The consistent decrease in $\text{PSNR}$ as high-$\text{EIG}$ regions are retained confirms that higher EIG marks lower-quality rendering.}
    \label{fig:EIG_metrics_trend}
\end{figure}
\\
\textbf{Pixel-wise Expected Information Gain with 3DGS.} Given a training set $D_{train}$ containing views $X_{train}$ (camera parameters) and corresponding image sequences $Y_{train}$, the differentiable renderer $\mathcal{F}$ of 3DGS is optimized over parameters $\mathbf{\omega}$. The goal is to obtain the point estimate $\mathbf{\omega}^{*}$ that minimizes the negative log-likelihood between rendered images $\hat{Y}_i^{train}=\mathcal{F}(X_i^{train}, \mathbf{\omega})$ and real observations $Y_i^{train}$. Assuming Gaussian observation noise, this is equivalent to minimizing the reconstruction error:

\vspace{-10pt}
\begin{equation}
\mathbf{\omega}^{*} = \arg \min_{\mathbf{\omega}} \sum_{(X_i, Y_i) \in \mathcal{D}_{train}} \| Y_i^{train} - \mathcal{F}(X_i^{train}, \mathbf{\omega}) \|_2^2
\label{eq:optimization_objective}
\end{equation}

\begin{figure*}[htbp]
    \centering
    \includegraphics[width=\linewidth]{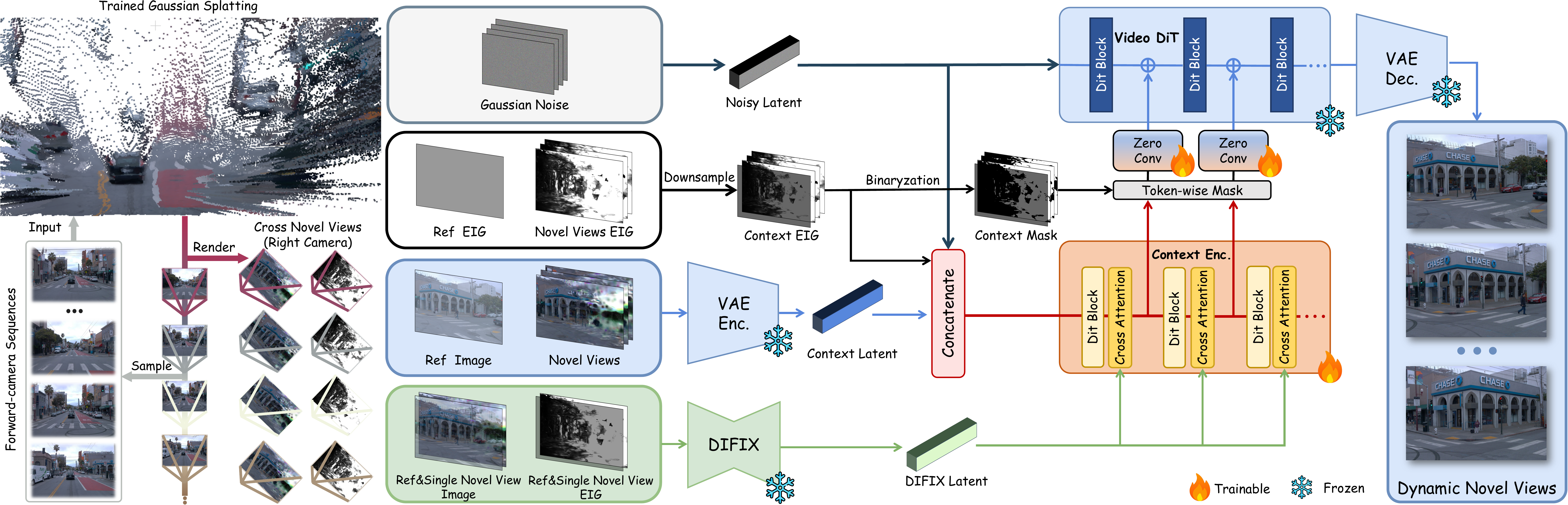}
    \caption{\textbf{Overview of EIGent.}
    \emph{Data:} Cross-view pairing: a forward-camera–trained 3DGS renders right-front views to produce artifact-prone novel-view renders and per-pixel EIG (Alg.~\ref{alg:EIG_compute}), temporally aligned with real right-front videos.
    \emph{Architecture:} EIGent is a dual-branch model with coarse-to-fine EIG guidance: downsampled $E$, noise latent $L_N$, and VAE latent $L$ feed a lightweight context encoder $\mathcal{G}$; a mask $M$ suppresses high–EIG regions. Via cross-attention with a \textsc{DIFIX} branch, cues are injected into a pretrained DiT backbone, enabling EIG-aware controllable repair and foreground spatio-temporal consistency.
    }
    \label{fig:eigent-pipline}
\end{figure*}

To quantify the uncertainty of $\mathbf{\omega}^{*}$ and estimate potential information gain from new observations, we apply the Laplace approximation~\cite{daxberger2021laplace}, modeling the posterior with a Gaussian distribution $\mathbf{\Omega}$:

\vspace{-10pt}
\begin{equation}
\mathbf{\Omega} \approx \mathcal{N}(\mathbf{\omega}^{*}, H^{\prime\prime}[\mathbf{\omega}^{*}]^{-1}),
\label{eq:laplace_approximation}
\end{equation}
where $H^{\prime\prime}[\mathbf{\omega}^{*}]$ is the Hessian of the negative log-likelihood at $\mathbf{\omega}^{*}$. The expectation of this Hessian under the predictive distribution corresponds to the Fisher information, quantifying how strongly the observations constrain the parameters. FisherRF~\cite{jiang2024fisherrf} treats the Fisher information as an uncertainty proxy, integrating it into the $\alpha$-blending process to compute pixel-level uncertainty at training views, but its application is limited to the observed training data.

For novel views $X_{NVS}$, the optimal 3DGS parameters $\mathbf{\omega}^{*}$ render novel-view sequences $Y_{NVS}$. The corresponding Expected Information Gain, defined as $\text{EIG}(\mathbf{\Omega}; Y_{i}^{NVS}|X_{i}^{NVS})$, is the difference between the prior entropy of $\mathbf{\Omega}$ and the expected posterior entropy after observing $Y_{i}^{NVS}$~\cite{houlsby2011bayesian}:

\vspace{-10pt}
\begin{equation}
\text{EIG} = \mathbb{H}[\mathbf{\Omega}] - \mathbb{E}_{p(Y_{i}|X_{i})}[\mathbb{H}[\mathbf{\Omega}|Y_{i}^{NVS},X_{i}^{NVS}]].
\label{eq:eig_definition}
\end{equation}
where $\mathbb{H}[\cdot]$ denotes the differential entropy of the distribution, and $\mathbb{E}_{p(\cdot)}[\cdot]$ denotes the expectation with respect to the predictive distribution $p(Y_{i}^{NVS}|X_{i}^{NVS})$.

Leveraging the Laplace approximation and properties of the Fisher information, this quantity can be efficiently estimated. To obtain a compact and computable upper bound, we apply the inequality $\log \det(A+I_d) \le \text{tr}(A)$ (Lemma 5.1)~\cite{kirsch2022unifying} and exploit the additive property of Fisher information, leading to the trace form\footnote{Please refer to the supplemental material for more details.\label{sup}}:
\vspace{-10pt}
\begin{equation}
\text{EIG} \le \frac{1}{2} \sum_{i} \text{tr}\left(H^{\prime\prime}[Y_{i}^{NVS}|X_{i}^{NVS},\mathbf{\omega}^{*}]H^{\prime\prime}[\mathbf{\omega}^{*}]^{-1}\right).
\label{eq:eig_trace_final}
\end{equation}
\vspace{-10pt}

While \cref{eq:eig_trace_final} yields the aggregate EIG for a full novel view, EIGent requires pixel-level granularity to guide local edits precisely. We extend EIG to each pixel by accumulating the Fisher information contributions of Gaussians intersected along each rendering ray. \cref{alg:EIG_compute} outlines the computation process: accumulating global Fisher information during training, and computing and mapping pixel-wise EIG for novel views. Following the methodology of BayesRays~\cite{goli2024bayes}, and as illustrated in \cref{fig:EIG_metrics_trend}, our cross-camera evaluations on the Waymo dataset~\cite{sun2020scalability} demonstrate that this pixel-level EIG strongly correlates with novel view synthesis quality\footref{sup}.
\input{algorithms/EIG_computation}

\subsection{EIGent: EIG Guided Dual-Branch Controllable Generation}
\label{sec:3.2}
\textbf{Dataset Generation.} To train the EIG-guided video restoration task, we construct datasets following the cross-camera referencing strategy of~\cite{wu2025difix3d+}. As illustrated in \cref{fig:eigent-pipline}, we first train a street-scene 3DGS model on forward-camera sequences, then render it from right-front view. This process simultaneously produces novel-view renderings and corresponding EIG maps computed via \cref{alg:EIG_compute}, which are paired with real observations to form triplets for the restoration task. To ensure data validity, we filter the raw sequences. We remove nearly stationary clips and compute inter-camera overlap (based on FOV) to guarantee sufficient reconstruction signals and adequate cross-view coverage. Furthermore, we mitigate EIG estimation distortion caused by large floaters near unseen regions by imposing explicit scale constraints on 3D Gaussians, confining artifacts within controllable bounds.\\
\textbf{EIGent architecture.}
$\text{EIGent}$ employs $\text{EIG}$ as a spatial prior for controllable restoration, providing interpretable, pixel-wise priorities:
\begin{itemize}
    \item \textbf{High-gain regions.} Areas of low rendering quality or missing information that require focused restoration and content generation.
    \item \textbf{Low-gain regions.} Backgrounds with reliable content where the model should preserve original structures.
\end{itemize}
To incorporate this prior effectively, we design a dual-branch control architecture, shown in \cref{fig:eigent-pipline}. A lightweight EIG-guided context encoder operates alongside a pretrained $\text{DiT}$ backbone, 

decoupling stable background preservation from temporally consistent foreground generation.

Given an input video $V$, a VAE encoder $\mathcal{E}$ maps it to a latent $L = \mathcal{E}(V)$, and the pixel-wise EIG map is downsampled to $E$. The multi-scale guidance strategy fuses these signals through EIG-guided context injection:
\begin{equation}
\label{eq:eigent_context_injection}
\epsilon_{\theta}(z_t, t, C)_{k} = \epsilon{\theta}(z_t, t, C)_{k} + M \odot \mathcal{G}(L_N, L, E)_{k},
\end{equation}
where $\epsilon_{\theta}$ denotes the DiT denoiser, $z_t$ the noisy latent, $t$ the timestep, $C$ the conditioning input, $\odot$ the Hadamard product, $L_N$ the noise latent, and $k$ the feature layer index. $\mathcal{G}$ represents the lightweight EIG-guided context encoder, cloned from the first four layers of the pretrained DiT.

Furthermore, to enhance per-frame quality, we fuse external repair cues (e.g., DIFIX latent) with the context branch $\mathcal{G}$ via cross-attention, regulating the fusion with spatial weights from $E$ and a binary mask $M$ that filters regions of extreme uncertainty (e.g., EIG above a threshold) and selectively incorporates reliable cues across multiple scales. This dual control injects coarse spatial metadata via $E$ while ensuring only trustworthy, background-relevant information reaches the DiT backbone, preventing contamination of stable contexts.

Overall, this EIG-driven coarse-to-fine guidance and fusion strategy enables full exploitation of the DiT architecture, improving both perceptual quality and spatio-temporal consistency in restored video.
\begin{figure*}[htbp]
    \centering
    \includegraphics[width=\linewidth]{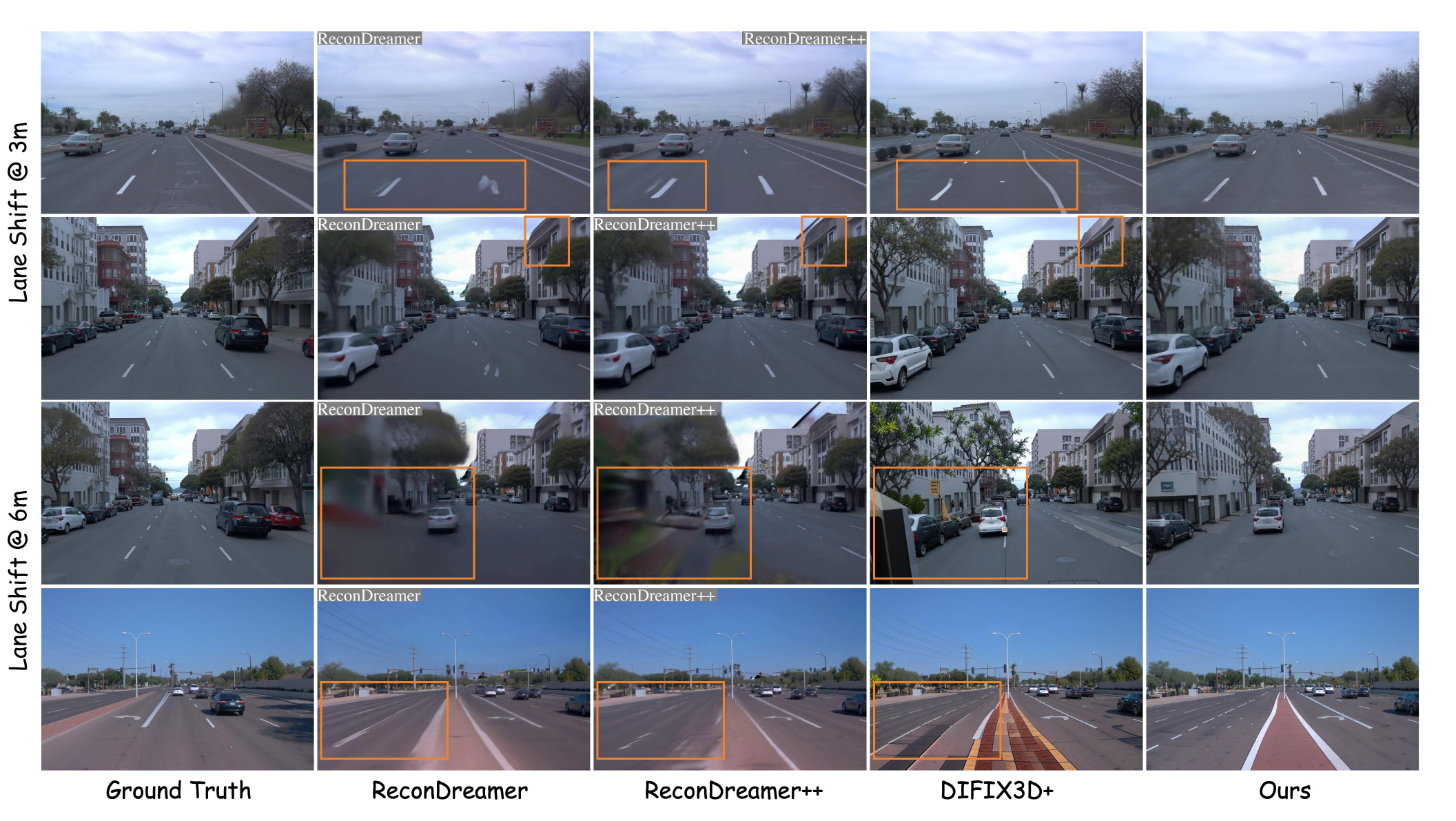}
    \caption{\textbf{Qualitative comparison on Waymo~\cite{sun2020scalability}.}
    Novel-view renderings for the same trajectory across representative methods~\cite{wang2024freevs,ni2025recondreamer,zhao2025recondreamer++,wu2025difix3d+}. 
    Orange boxes highlight regions where our approach yields noticeably better results.}
    \label{fig:metric_gen_contrast}
\end{figure*}
\subsection{Progressive EIG-Aware Diffusion-to-3DGS Knowledge Integration}
\label{sec:3.3}
To fully exploit the restoration capability of EIGent and inject high-quality generated content into 3DGS in an orderly and controllable manner, we adopt a progressive knowledge integration strategy inspired by prior work~\cite{ni2025recondreamer, wu2025difix3d+}. Different from existing approaches that rely on heuristics or view-level control, our key idea is to use \textbf{pixel-wise EIG} as the guiding signal, enabling finer-grained and more interpretable fusion.

Taking a standard street-scene 3DGS pipeline~\cite{yan2024street, chen2024omnire} as an example: the overall loss is composed of the original-trajectory term and the novel-trajectory term. The loss for the original trajectory is

\begin{equation} \label{eq:train_loss}
\mathcal{L}_{ori}(\omega) = \lambda_r \mathcal{L}{1}^{ori} + (1-\lambda_r)\mathcal{L}_{SSIM}^{ori} + \lambda_d\mathcal{L}_{depth}^{ori},
\end{equation}
where $\mathcal{L}_{1}$ and $\mathcal{L}_{SSIM}$ denote the L1 and SSIM losses between rendered images and ground-truth views, $\mathcal{L}_{depth}$ is the depth supervision against sparse LiDAR, and $\lambda_r,\lambda_d$ are the corresponding weights.

To incorporate the restored novel views into the optimization, we introduce a novel-view loss $\mathcal{L}_{\text{novel}}(\omega)$ with two components:
\begin{itemize}
    \item \textbf{EIG pixel-wise weighting.} The normalized EIG map is used as a pixel-level weight matrix $\lambda_{\text{EIG}}$ to modulate the image loss, so that 3DGS focuses optimization on the regions with the highest information gain (i.e., the most under-constrained areas):
    \begin{equation} \label{eq:novel_loss_image}
    \mathcal{L}_{\text{img}}^{\text{novel}} = \lambda_{\text{EIG}} \odot \big(\lambda_r \mathcal{L}_{1}^{\text{novel}} + (1-\lambda_r)\mathcal{L}_{\text{SSIM}}^{\text{novel}}\big)
    \end{equation}
    \item \textbf{Sparse depth supervision.} We further employ point-cloud projections aggregated from neighboring frames as sparse depth supervision $\mathcal{L}_{\text{depth}}^{\text{novel}}$ to preserve geometric consistency in novel views.
\end{itemize}
The novel-trajectory loss is thus
\begin{equation}
\mathcal{L}_{\text{novel}}(\omega) = \mathcal{L}_{\text{img}}^{\text{novel}} + \lambda_d \mathcal{L}_{\text{depth}}^{\text{novel}},
\label{novel loss}
\end{equation}
and is used to fine-tune the existing 3DGS model. This fine-tuning relies on EIGent-restored views in the early stage to prioritize spatial structure and cross-frame coherence, followed by DIFIX3D+ refined views once the expansion reaches its maximum range and stabilizes.
\begin{table*}[t]
\centering
\setlength{\tabcolsep}{3pt} 
\resizebox{0.9\linewidth}{!}{
\begin{tabular}{@{}l ccc ccc ccc@{}}
\toprule
\multirow{2}{*}{Method} & \multicolumn{3}{c}{Extra Condition} & \multicolumn{3}{c}{Lane Shift @ 3m} & \multicolumn{3}{c}{Lane Shift @ 6m} \\
\cmidrule(lr){2-4} \cmidrule(lr){5-7} \cmidrule(lr){8-10}
& LiDAR & Box & HDMap & NTA-IoU $\uparrow$ & NTL-IoU $\uparrow$ & FID$\downarrow$ & NTA-IoU $\uparrow$ & NTL-IoU $\uparrow$ & FID$\downarrow$ \\
\midrule
OmniRe ~\cite{chen2024omnire}  & $\times$ & $\times$ & $\times$ & 0.424 & 51.73 & 188.42 & 0.423 & 49.08 & 191.00 \\
FreeVS ~\cite{wang2024freevs} & \checkmark & $\times$ & $\times$ & 0.505 & 56.84 & 104.23 & 0.465 & 55.37 & 121.44 \\
\midrule
ReconDreamer ~\cite{ni2025recondreamer}  & $\times$ &\checkmark & \checkmark & 0.539 & 54.58 & 93.56 & 0.467 & 52.58 & 149.19 \\
ReconDreamer++~\cite{zhao2025recondreamer++}\textbf{*} & $\times$ & \checkmark & \checkmark & 0.572 & \underline{57.06} & \underline{72.02} & 0.489 & \textbf{56.57} & \underline{111.92} \\
DIFIX3D+ ~\cite{wu2025difix3d+}& $\times$ & $\times$ & $\times$ & \underline{0.578} & 56.94 & 84.12 & \underline{0.504} & 53.77 & 120.24 \\
\midrule
FaithFusion & $\times$ & $\times$ & $\times$ & \textbf{0.581} & \textbf{57.67} & \textbf{71.51} & \textbf{0.517} & \underline{55.78} & \textbf{107.47} \\
\bottomrule
\end{tabular}}
\caption{
Comparison of different lane shifts on the Waymo dataset~\cite{sun2020scalability}, highlighting key methodology requirements. Extra Condition indicates the reliance on additional data injected as a condition to guide the synthesis process (e.g., LiDAR, 3D boxes, HDMap). $\mathbf{*}$ denotes that the method requires significant architectural or geometrical modifications, including decomposed modeling and new trajectory field.
}
\label{tab:waymo}
\end{table*}

\section{Experiments}
\label{sec:experiments}
\subsection{Experiment Setup}
\textbf{3DGS Training and Evaluation Protocol.}
We conduct our experiments on the Waymo dataset~\cite{sun2020scalability}, strictly following the experimental protocol established by ReconDreamer~\cite{ni2025recondreamer}. Specifically, the 3DGS model is trained on $\text{8}$ distinct clips ($\text{40}$ frames per clip) using only the forward-camera data. During evaluation, we assess the cross-lane rendering quality. The training trajectory is expanded progressively by $\text{1.0 m}$ every $\text{2,000}$ iterations starting from step $\text{3,000}$ for synthesizing novel viewpoints.\\
\textbf{EIGent Model Fine-Tuning and Data Preparation.}
The video generation components, the external DIFIX model and the dual-branch video generator within EIGent, are fine-tuned using a dedicated dataset. We preprocess the first $\text{200}$ training clips of the Waymo dataset~\cite{sun2020scalability} for this purpose. Applying the data filtering strategy detailed in \cref{sec:3.2}, this process yields $\text{936}$ video triplets.
The image-generation branch (DIFIX) is fine-tuned using LoRA under the default DIFIX3D+~\cite{wu2025difix3d+} configuration. Meanwhile, for the video branch (EIGent), we freeze the base diffusion model (CogVideo-5B-I2V~\cite{yang2024cogvideox}) and train only the context encoder for $\text{14,336}$ steps with a learning rate of $1\times10^{-5}$.\\
\textbf{Baselines.}
To comprehensively evaluate \emph{FaithFusion}, we integrate it into the general 3DGS reconstruction framework OmniRe~\cite{chen2024omnire} and compare it with FreeVS~\cite{wang2024freevs} as a representative generation method, as well as three fusion-based novel view synthesis methods, including ReconDreamer~\cite{ni2025recondreamer}, ReconDreamer++~\cite{zhao2025recondreamer++}, and DIFIX3D+~\cite{wu2025difix3d+}.\\
\textbf{Evaluation Metrics.}
Following DriveDreamer4D~\cite{zhao2025drivedreamer4d}, we report Novel Trajectory Agent IoU (NTA-IoU), Novel Trajectory Lane IoU (NTL-IoU), and FID as the primary evaluation metrics. We also introduce two EIG-partitioned metrics in our ablation study: FID-UCR, assessing Under-Constrained Regions (UCR), and FID-HPR, for High-Confidence Regions (HPR).

\subsection{Comparison with State-of-the-Art Methods}
Comparison among representative SOTA methods is shown in \cref{tab:waymo}. For a fair comparison, we crop the outputs of FreeVS~\cite{wang2024freevs} to exclude regions without LiDAR coverage when computing metrics. Benefiting from the plug-and-play nature of DIFIX3D+~\cite{wu2025difix3d+}, we integrate it into OmniRe~\cite{chen2024omnire} using the official gsplat~\cite{ye2025gsplat} interface. 

At a lane shift of 3 meters, \emph{FaithFusion} approaches the best IoU scores and achieves the lowest FID (71.51), demonstrating strong 3D semantic stability and visual generalization. This robustness stems from EIG's dual role in the video-generation and DIFIX3D+~\cite{wu2025difix3d+} restoration stages, and from the differentiated constraints of the EIG-weighted reconstruction loss. High-confidence regions that are co-visible with the original trajectory are refined for details while maintaining structural coherence, whereas under-constrained regions are guided to generate plausible geometry and semantics. As a result, progressive updates yield stable and high-quality novel-view generation even at moderate trajectory deviations. Visualization results (upper half of \cref{fig:metric_gen_contrast}) show that road structures from the original view are well preserved under EIG guidance; in synthesized regions, the EIG map clarifies repair needs, enabling semantically correct completion (e.g., building facades) and reducing 3D semantic inconsistencies.
\begin{figure*}[!htbp]
    \centering
    \includegraphics[width=\linewidth]{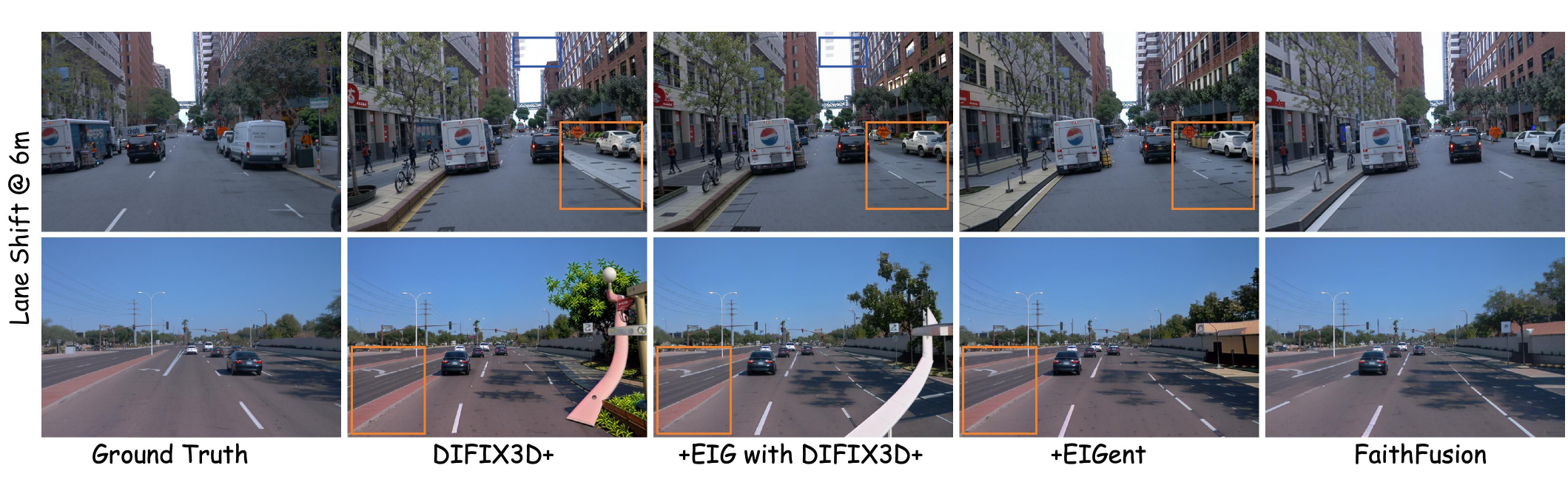}
    \caption{\textbf{Ablation overview.}
    We incrementally integrate $\text{EIG}$-guided components into the $\text{OmniRe}$ baseline.  The results highlight the incremental contributions of $\text{EIG}$ guidance in resolving over-restoration and geometric drift by acting as a unified pixel-wise editing policy.}
    \label{fig:paper_ablation}
\end{figure*}

At a lane shift of 6 meters, most methods suffer severe performance degradation due to accumulated errors, whereas \emph{FaithFusion} maintains stable performance (NTA-IoU: 0.517, NTL-IoU: 55.78, FID: 107.47). 
Here, EIG guides the model to focus on critical unseen structures during generation to ensure geometric fidelity, while constraining high-confidence regions during repair to seamlessly integrate new content with existing areas, thus achieving a balance between global coherence and fine-detail fidelity.
Visualization results (lower half of \cref{fig:metric_gen_contrast}) illustrate these effects: while competing methods often produce blurred or semantically inconsistent results, EIGent—guided explicitly by EIG and enhanced by robust video generation—precisely localizes and restores missing regions. Our results exhibit state-of-the-art global coherence, avoiding spurious deformations such as the erroneous ground bending in the third row, and preventing unrealistic artifacts often observed when using DIFIX3D+~\cite{wu2025difix3d+} alone.

\begin{table}[t] 
\centering
\setlength{\tabcolsep}{2pt} 
\resizebox{0.8\linewidth}{!}{\begin{tabular}{@{}l ccc@{}} 
\toprule
\multirow{2}{*}{Method} & \multicolumn{3}{c}{FID $\downarrow$} \\ 
\cmidrule(lr){2-4} 
& Total & UCR & HPR \\ 
\midrule
\makecell[l]{DIFIX3D+ \\ (Baseline)} & 120.24 & 147.97 & 152.66 \\ 
+ EIG Guided DIFIX3D+ & 119.01 & 143.80 & 149.82 \\
++ EIGent Dual-Stage Fusion & 113.94 & 137.58 & 153.69 \\
\midrule
\makecell[l]{\textbf{+++ EIG Recon} \\ \textbf{(Full FaithFusion)}} & \textbf{107.47} & \textbf{137.02} & \textbf{147.75} \\
\bottomrule
\end{tabular}}
\caption{\textbf{Ablation Study: Incremental Contributions of FaithFusion's Core Components.} Results on the most challenging 6-meter lateral-shift novel-view synthesis task, showing the gain from sequentially adding the three proposed EIG-guided modules to the $\text{DIFIX3D+}$ baseline.}

\label{tab:ablation}
\end{table}
\subsection{Ablation Study}
\label{sec:ablation}
\begin{figure}[!htbp]
    \centering
    \includegraphics[width=\linewidth]{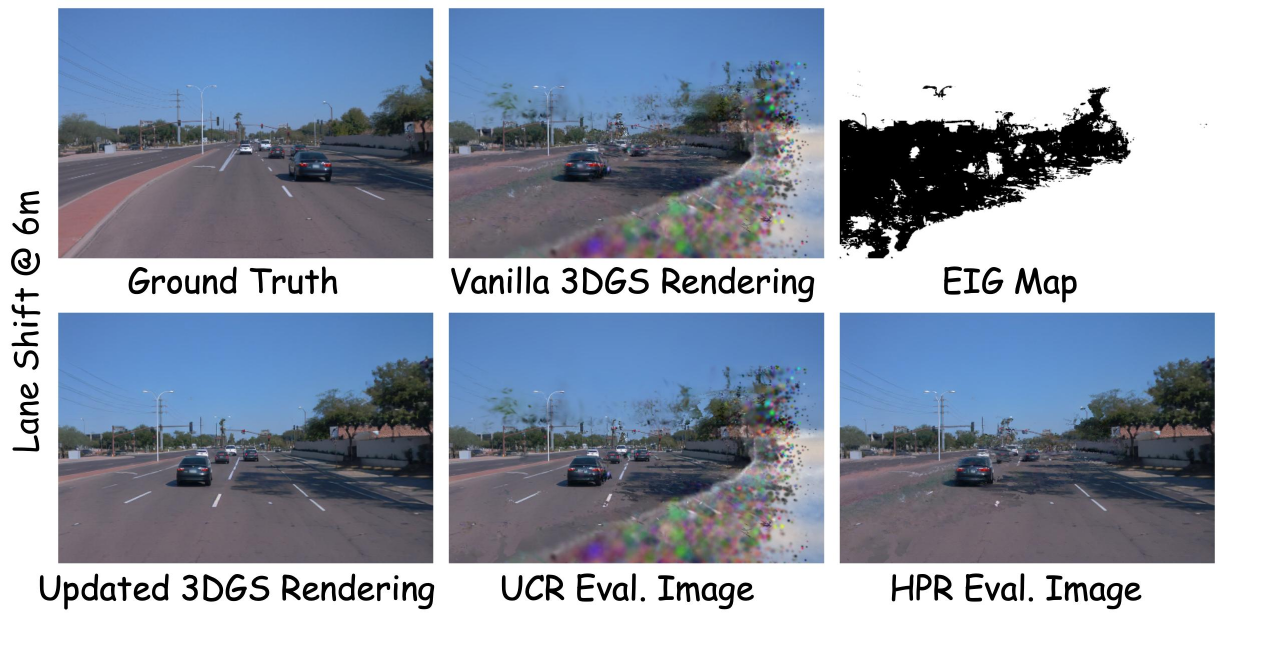}
    \caption{\textbf{Region partition for EIG-based evaluation.}
    We use EIG as a proxy for rendering quality. With a threshold $\tau\!=\!0.4$, renderings are partitioned into UCR and HPR, and we report region-specific metrics: $\text{FID-UCR}$ and $\text{FID-HPR}$.}
    \label{fig:ablation_judge}
\end{figure}

We conduct comprehensive ablations to quantify the contribution of each component in \emph{FaithFusion}. We argue that traditional global FID is insufficient to accurately measure fine-grained performance differences across regions of varying confidence. Given that pixel-wise EIG effectively reflects local rendering quality, we introduce two EIG-partitioned metrics to distinguish performance across uncertainty levels. Specifically, we define Under-Constrained Regions (UCR) and High-Confidence Regions (HPR), with the corresponding metrics being FID-UCR and FID-HPR. These regions are defined by the EIG threshold detailed in \cref{fig:ablation_judge}. During evaluation, the non-evaluated region is filled with vanilla 3DGS renderings to ensure complete inputs.

Overall, the improvements of our full system stem from three complementary components operating at different stages:
As shown in \cref{tab:ablation}, introducing EIG guidance consistently reduces FID, with a total drop of about 1.23. Compared with pure DIFIX3D+~\cite{wu2025difix3d+}, EIG focuses the repair on low-EIG (well-reconstructed) regions and suppresses unnecessary hallucinations, thereby improving overall 3DGS performance. The visualization results this trend: DIFIX3D+~\cite{wu2025difix3d+} with EIG guidance alleviates semantic mismatches in low-EIG areas, as visually confirmed by \cref{fig:paper_ablation}, although high-EIG regions may still exhibit deviations due to the lack of strong priors.

Adding EIGent further enhances generation quality across both partitioned regions, reducing the overall FID by 5.07 and lowering FID-UCR by 6.22. This indicates that EIGent effectively compensates for weaknesses in novel-view synthesis and makes the outputs closer to real images. 

A slight increase in FID-HPR is observed, which mainly comes from enforcing temporal consistency in video diffusion, where stronger consistency can dampen fine-grained appearance details. Nonetheless, the visualization results, as shown in the lower half of \cref{fig:paper_ablation}, show that EIGent produces semantically more coherent and plausible content, especially in regions that require completion or semantic reasoning.

Finally, the progressive EIG-aware diffusion-to-GS integration reduces the total FID to \textbf{107.47}, improving by 12.77 over the baseline. FID-UCR and FID-HPR decrease by 10.95 and 4.91, respectively. This strategy retains EIGent’s benefits while preventing over-restoration in low-EIG regions, ensuring structural and appearance consistency. Visualization results further confirm these gains, showing a balanced trade-off between detail restoration and global coherence.

\vspace{-10pt}
\section{Conclusion}
\label{sec:conclusion}

We introduce \emph{FaithFusion}, a 3DGS–diffusion fusion paradigm driven by pixel-wise Expected Information Gain (EIG), which unifies faithful reconstruction and controllable generation by converting heuristic editing decisions into an information-theoretic quantity. This cross-modal EIG guidance employs EIG as a spatial weight for content suppression on the generation side and as a loss weight for selective knowledge distillation on the reconstruction side, offering strong generality and interpretability. Systematic experiments on the Waymo dataset~\cite{sun2020scalability} demonstrate that \emph{FaithFusion} significantly improves spatio-temporal consistency and perceptual quality under large viewpoint shifts, achieving state-of-the-art results across major metrics.\\\\
\textbf{Limitations and Future Work.} While EIG effectively slows the accumulation of errors inherent in the 3DGS–diffusion fusion paradigm, the issue is not fully eliminated; this suggests that customizing the 3DGS model architecture could be key to further reduction. Furthermore, EIG is widely utilized in active exploration and mapping, and its incorporation into \emph{FaithFusion} provides a natural bridge for future work, enabling active mapping strategies to significantly enhance overall efficiency.

{\small
\bibliographystyle{ieeenat_fullname}
\bibliography{main}
}

\input{X_suppl}
\end{document}

%% file: algorithms/EIG_computation.tex
\begin{algorithm}[h] 
    \caption{Pixel-wise Expected Information Gain Computation in 3DGS}
    \label{alg:EIG_compute}
    \begin{algorithmic}[1] 
        \Require Trained 3DGS model parameters $\mathbf{\omega}^{*}$, Total number of 3D Gaussians $N$, Train views $X_{train}$, Novel views $X_{NVS}$, Differentiable rasterization $\mathcal{F}$, Hessian computation function(Laplace diag, per-Gaussian) $\mathcal{H}$
        \Ensure \raggedright Pixel-wise Expected Information Gain $EIG_{NVS}$            
        \State $H^{\prime\prime}[\mathbf{\omega}^{*}] \leftarrow \mathbf{0} \in \mathbb{R}^{N}$
        \For{each train view $X_i^{train}$ in $X_{train}$}
            \State $H^{\prime\prime}[\mathbf{\omega}^{*}] \leftarrow H^{\prime\prime}[\mathbf{\omega}^{*}] + \mathcal{H}[\mathcal{F}(X_i^{train}, \mathbf{\omega}^{*})]$
        \EndFor
        \For{each novel view $X_j^{NVS}$ in $X_{NVS}$}
            \State $H^{\prime\prime}[Y_j^{NVS}|X_j^{NVS},\mathbf{\omega}^{*}] \leftarrow \mathcal{H}[\mathcal{F}(X_j^{NVS}, \mathbf{\omega}^{*})]$
            \State $EIG_j^{GS} \leftarrow H^{\prime\prime}[Y_j^{NVS}|X_j^{NVS},\mathbf{\omega}^{*}] \odot (H^{\prime\prime}[\mathbf{\omega}^{*}])^{-1}$
            \State $EIG_j^{NVS} = \mathcal{F}((X_j^{NVS},EIG_j^{GS}),\mathbf{\omega}^{*})$
        \EndFor
    \end{algorithmic}
\end{algorithm}

%% file: X_suppl.tex
\clearpage
\setcounter{page}{1}
\appendix
\renewcommand{\thesection}{\Alph{section}}
\setcounter{section}{0}
\renewcommand{\thefigure}{S\arabic{figure}}
\renewcommand{\thetable}{S\arabic{table}}
\renewcommand{\theequation}{S\arabic{equation}}
\setcounter{figure}{0}
\setcounter{table}{0}
\setcounter{equation}{0}

\maketitlesupplementary

\section{Additional Implementation Details}
\label{sec:Additional Implementation Details}
\subsection{Driving Scene Representation}
\label{sec:sup_scene_rep}
FaithFusion's scene representation is built upon the decoupled 3D Gaussian Splatting (3DGS) structure widely adopted for driving scenes~\cite{yan2024street, chen2024omnire, zhao2025recondreamer++}. This representation is explicitly decomposed into three core components: a \emph{Static Background} ($\mathcal{G}^\text{bg}$), \emph{Dynamic Rigid Objects} ($\bar{\mathcal{G}}_v^\text{rigid}$ in local canonical space), and a \emph{Sky Model} ($C_{\text{sky}}$), to precisely model geometry and motion through distinct semantic representations.\\
\textbf{Static Background.}
The static background component is represented by a set of static Gaussian primitives $\mathcal{G}^\text{bg}$. These Gaussians are defined and optimized directly in the global world coordinate system, and all their attributes remain invariant over time. \\
\textbf{Rigid Body Representation and Transformation.}
Gaussians belonging to a rigid object $v$ are defined in its local canonical space. The set of Gaussians in this space, $\bar{\mathcal{G}}_v^\text{rigid}$, do not change their internal attributes (mean $\bar{\boldsymbol{\mu}}$, rotation $\bar{\mathbf{q}}$, scale $\bar{\mathbf{s}}$, opacity $\bar{o}$, and SH coefficients $\bar{\mathbf{c}}$) over time $t$. The object's motion is captured entirely by an external rigid transformation $\mathbf{T}_v(t) \in \text{SE}(3)$, which transforms the Gaussians into world space $\mathcal{G}^\text{rigid}_v(t)$:
\begin{equation}
    \mathcal{G}^\text{rigid}_v(t) = \mathbf{T}_v(t) \otimes \bar{\mathcal{G}}^\text{rigid}_v.
\end{equation}

The transformation operator $\otimes$ specifically updates the mean position $\boldsymbol{\mu}(t)$ and rotation $\mathbf{q}(t)$ of the Gaussians when moved to the world coordinate system, while other attributes (scale, opacity, and SH coefficients) remain unchanged. We decompose the rigid pose as $\mathbf{T}_v(t)=(\mathbf{R}_v(t), \mathbf{t}_v(t))$, and the world-space mean position $\boldsymbol{\mu}(t)$ is obtained by:
\begin{equation}
    \boldsymbol{\mu}(t) = \mathbf{R}_v(t) \bar{\boldsymbol{\mu}} + \mathbf{t}_v(t),
\end{equation}
the rotation $\mathbf{q}(t)$ is updated by composing the object's rotational component $\mathbf{R}_v(t)$ with the canonical rotation $\bar{\mathbf{q}}$:
\begin{equation}
    \mathbf{q}(t) = \mathrm{Rot}(\mathbf{R}_v(t), \bar{\mathbf{q}}),
\end{equation}
where $\mathrm{Rot}(\cdot)$ denotes rotating the quaternion $\bar{\mathbf{q}}$ by the rotation matrix $\mathbf{R}_v(t)$. In this manner, the motion of dynamic objects is accurately modeled, ensuring geometric consistency across the time sequence.\\
\textbf{Sky Model Compositing.}
The sky model is treated as a separate optimizable environmental texture map $C_{\text{sky}}$ to fit large-scale appearance. The final pixel color $C$ is obtained by $\alpha$-blending the rendered Gaussian image $C_{\mathcal{G}}$ with the sky image $C_{\text{sky}}$, where $C_{\mathcal{G}}$ is rendered from all Gaussians ($\mathcal{G}^\text{bg}$ and $\{\mathcal{G}^\text{rigid}_v\}$):
\begin{equation}
    C = C_{\mathcal{G}} + (1 - O_{\mathcal{G}}) C_{\text{sky}},
\end{equation}
where $O_{\mathcal{G}}$ is the rendered opacity mask accumulated from all Gaussian primitives. This strategy addresses the challenge of reconstructing unbounded distant scenes.

\subsection{Evaluation Metrics}
We conduct comprehensive quantitative evaluations following the protocol established by DriveDreamer4D~\cite{zhao2025drivedreamer4d}, utilizing metrics that jointly assess the 3DGS novel view synthesis capability and the resulting spatiotemporal consistency under complex conditional shifts (e.g., lane shifts).\\
\textbf{Spatiotemporal Coherence Metrics. ($\uparrow$)}
To rigorously evaluate the coherence of dynamic elements and static scene structure, we employ two core metrics. Both quantify accuracy by comparing features detected in the rendered image with ground truth features that are geometrically projected from the original 3D scene onto the new trajectory view.

\begin{itemize}
    \item \textbf{Novel Trajectory Agent IoU ($\text{NTA-IoU}$):} Measures the spatiotemporal accuracy of foreground dynamic agents (vehicles). Its computation involves detecting 2D bounding boxes on the rendered frames and comparing them to the projected ground-truth 3D bounding boxes. High $\text{NTA-IoU}$ ensures accurate agent placement and adherence to the underlying 3D structure, leading to precise corrections in under-constrained regions.
    \item \textbf{Novel Trajectory Lane IoU ($\text{NTL-IoU}$):} Measures the geometric fidelity and spatiotemporal coherence of background lane lines. By comparing lane lines detected in the synthesized image against projected ground truth (often derived from the HDMap), it specifically verifies the integrity of the environment's static geometry. This metric reflects minimal disturbance to the original scene structure and guarantees environmental consistency.
\end{itemize}
\textbf{Perceptual Quality Metric. ($\downarrow$)}
We use Fréchet Inception Distance ($\text{FID}$)~\cite{heusel2017gans} to evaluate the overall visual realism and distributional quality of the 3DGS rendered novel view frames. $\text{FID}$ calculates the distance between two multivariate Gaussian distributions fitted to the deep feature representations (from an Inception network) of generated frames and real frames. This score reflects the distribution-level similarity in a high-level perceptual space. A lower $\text{FID}$ score indicates superior visual quality and consistent behavior across diverse viewpoints.

\onecolumn
\subsection{Expected Information Gain Derivations}
\label{sec:eig_derivation}

We provide the detailed derivation for the Expected Information Gain (EIG) approximation utilized in the main paper (Equation \ref{eq:eig_definition}). This derivation strictly follows the unified framework for Bayesian optimal experimental design and information-theoretic approximations \cite{kirsch2022unifying, houlsby2011bayesian}.\\
\textbf{Motivation.} The analytical computation of the EIG definition in Equation \ref{eq:eig_definition} is intractable, requiring evaluation of complex posterior parameter distributions and expectations over the observation space. To obtain a highly efficient and differentiable acquisition function for 3DGS, our goal is to derive a \emph{computable upper bound} of the EIG. This is achieved by combining the \emph{Laplace approximation} (to simplify the entropy terms, Prop. 3.2/3.5 in \cite{kirsch2022unifying}) and the \emph{log-determinant inequality} (to obtain the final trace form upper bound, Lemma 5.1 in \cite{kirsch2022unifying}).\\
\textbf{EIG Definition.}
The EIG quantifies the \emph{predicted reduction in uncertainty of 3DGS model parameters $\mathbf{\Omega}$} if a new observation ($Y_{NVS}^{\text{gt}}$) at the novel view $X_{NVS}$ is acquired. Following the mutual information definition of EIG in \cite{kirsch2022unifying} (Section 5.1), this uncertainty reduction is formally the mutual information between $\mathbf{\Omega}$ and $Y_{NVS}^{\text{gt}}$ (conditioned on $X_{NVS}$):
\vspace{-5pt}
\begin{equation}
  \text{EIG} = I\left[\mathbf{\Omega}; Y_{NVS}^{\text{gt}} \mid X_{NVS}\right] = \mathbb{H}[\mathbf{\Omega}] - \mathbb{E}_{p(Y_{NVS}^{\text{gt}} \mid X_{NVS})}\left[\mathbb{H}\left[\mathbf{\Omega} \mid Y_{NVS}^{\text{gt}}, X_{NVS}\right]\right],
  \label{eq:total_EIG_gt_Definition}
\end{equation}

where $\mathbb{H}[\mathbf{\Omega}]$ denotes the \emph{prior entropy}, and $\mathbb{H}[\mathbf{\Omega} \mid Y_{NVS}^{\text{gt}}, X_{NVS}]$ denotes the \emph{posterior entropy}.\\
\textbf{\textcolor{red!70!black}{! Note on Observation Index $i$ and EIG Decomposition.}} The index $i$ in the main paper's equations is used to denote different scopes of observation:
\begin{itemize}
    \item In the optimization objective (Eq. \ref{eq:optimization_objective}), $i$ denotes an individual training view.
    \item In the EIG definition (Eq. \ref{eq:eig_definition}), $i$ denotes an individual view $Y_i^{NVS}$ within the novel view sequence $Y_{NVS}$. This formula calculates the information gain from a single such view.
\end{itemize}
The full EIG for the entire novel view sequence is obtained via view additivity. The final computable bound (Eq. \ref{eq:eig_trace_final}) is then derived by applying Fisher information additivity principle (Prop. 4.2 in \cite{kirsch2022unifying}) across all pixels $j$ in every view $i$. Therefore, the summation index $i$ in the final trace form (Eq. \ref{eq:eig_trace_final}) is implicitly a flattened sum over all pixels in novel view sequence.\\
\textbf{Laplace Proxy Justification.}
In \cref{eq:total_EIG_gt_Definition}, $p(Y_{NVS}^{\text{gt}} \mid X_{NVS})$ is the true predictive distribution of real observations. We use the deterministic 3DGS rendered result $Y_{NVS}$ as a computationally tractable proxy for $Y_{NVS}^{\text{gt}}$. This is justified by the Laplace approximation (Prop. 3.2 in \cite{kirsch2022unifying}), which models 3DGS parameters $\mathbf{\Omega}$ as a Gaussian posterior around the MAP parameters $\mathbf{\omega}^*$ ($\mathbf{\Omega} \sim \mathcal{N}(\mathbf{\omega}^*, (H''[\mathbf{\omega}^*])^{-1})$). Here, $H''[\mathbf{\omega}^*]$ is the Hessian of the negative log-posterior, serving as the inverse prior covariance. Since $Y_{NVS}$ is a deterministic function of $\mathbf{\Omega}$, $Y_{NVS}$ (conditioned on $\mathbf{\omega}^*$) approximates $p(Y_{NVS}^{\text{gt}} \mid X_{NVS})$ for this Gaussian parameter posterior. \\
\textbf{Laplace Approximation of Entropy.}
To compute the entropy terms in \cref{eq:total_EIG_gt_Definition}, we apply the Gaussian differential entropy formula:$\mathbb{H}[\mathcal{N}(\mu, \boldsymbol{\Sigma})] = \frac{1}{2} \log \det(2\pi e \boldsymbol{\Sigma})$, where the covariance $\boldsymbol{\Sigma}$ is the inverse of the \emph{observed information matrix} (Hessian of the negative log-posterior, Prop. 3.2 in \cite{kirsch2022unifying}). For EIG, we distinguish two key observed information matrices:
\begin{itemize}
    \item \textbf{Prior observed information:} $H''[\mathbf{\omega}^*]$ (Hessian of the negative log-posterior of $\mathbf{\Omega}$ evaluated at $\mathbf{\omega}^*$, corresponding to $\mathbb{H}[\mathbf{\Omega}]$);
    \item \textbf{Posterior observed information:} $H''[\mathbf{\Omega} \mid Y_{NVS}] = H''[\mathbf{\omega}^*] + H''[Y_{NVS} \mid \mathbf{\omega}^*]$ (sum of prior information and novel-view information, via the information additivity principle in Prop. 4.2 of \cite{kirsch2022unifying}), where $H''[Y_{NVS} \mid \mathbf{\omega}^*]$ is the Hessian of the negative log-likelihood of $Y_{NVS}$ (conditioned on $\mathbf{\omega}^*$).
\end{itemize}
Substituting these Gaussian entropy approximations into \cref{eq:total_EIG_gt_Definition} yields:
\vspace{-5pt}
\begin{align}
  \text{EIG} &\approx \frac{1}{2} \log \det\left(2\pi e (H''[\mathbf{\omega}^*])^{-1}\right) - \mathbb{E}_{p(Y_{NVS} \mid X_{NVS}, \mathbf{\omega}^*)}\left[\frac{1}{2} \log \det\left(2\pi e \left(H''[\mathbf{\omega}^*] + H''[Y_{NVS} \mid \mathbf{\omega}^*]\right)^{-1}\right)\right] \\
  &= \frac{1}{2} \left[ \log \det\left((H''[\mathbf{\omega}^*])^{-1}\right) - \mathbb{E}_{p(Y_{NVS} \mid X_{NVS}, \mathbf{\omega}^*)}\left[\log \det\left(\left(H''[\mathbf{\omega}^*] + H''[Y_{NVS} \mid \mathbf{\omega}^*]\right)^{-1}\right)\right] \right] \\
  &= \frac{1}{2} \mathbb{E}_{p(Y_{NVS} \mid X_{NVS}, \mathbf{\omega}^*)}\left[ \log \det\left(H''[\mathbf{\omega}^*] + H''[Y_{NVS} \mid \mathbf{\omega}^*]\right) - \log \det\left(H''[\mathbf{\omega}^*]\right) \right] \\
  &= \frac{1}{2} \mathbb{E}_{p(Y_{NVS} \mid X_{NVS}, \mathbf{\omega}^*)}\left[ \log \det\left(\mathbf{I} + H''[Y_{NVS} \mid \mathbf{\omega}^*] (H''[\mathbf{\omega}^*])^{-1}\right) \right].
\label{eq:laplace_Approximation_of_Entropy}
\end{align}\\
\textbf{Trace Form Upper Bound and Pixel-Level Decomposition.}
We apply the log determinant inequality ($\log \det(\mathbf{I} + \mathbf{A}) \le \text{tr}(\mathbf{A})$) (Lemma 5.1 in \cite{kirsch2022unifying}) to \cref{eq:laplace_Approximation_of_Entropy}. By the linearity of the trace operator, and substituting the expectation of the novel-view Hessian $\mathbb{E}_{p(Y_{NVS} \mid X_{NVS}, \mathbf{\omega}^*)}\left[ H''[Y_{NVS} \mid \mathbf{\omega}^*] \right]$ with its Fisher information (Prop. 4.1 in \cite{kirsch2022unifying}), we get:
\vspace{-5pt}
\begin{align}
  \text{EIG} &\le \frac{1}{2} \mathbb{E}_{p(Y_{NVS} \mid X_{NVS}, \mathbf{\omega}^*)}\left[ \text{tr}\left( H''[Y_{NVS} \mid \mathbf{\omega}^*] (H''[\mathbf{\omega}^*])^{-1} \right) \right] \\
  &= \frac{1}{2} \text{tr}\left( \mathbb{E}_{p(Y_{NVS} \mid X_{NVS}, \mathbf{\omega}^*)}\left[ H''[Y_{NVS} \mid \mathbf{\omega}^*] \right] (H''[\mathbf{\omega}^*])^{-1} \right) \\
  &= \frac{1}{2} \text{tr}\left( H''[Y_{NVS} \mid X_{NVS}, \mathbf{\omega}^*] (H''[\mathbf{\omega}^*])^{-1} \right).
  \label{eq:eig_pixel_view}
\end{align}

Finally, leveraging the Fisher information additivity principle (Prop. 4.2 in \cite{kirsch2022unifying}), the total Fisher information of $Y_{NVS}$ is the sum of pixel-level Fisher information $H''[Y_{i,NVS} \mid X_{i,NVS}, \mathbf{\omega}^*]$ (one per pixel $i$). Substituting these pixel-wise Fisher information into \cref{eq:eig_pixel_view} yields the final trace-form approximation (main paper Equation \ref{eq:eig_trace_final}):
\vspace{-5pt}
\begin{equation}
   \text{EIG} \le \frac{1}{2} \sum_{i} \text{tr}\left( H''[Y_{i,NVS} \mid X_{i,NVS}, \mathbf{\omega}^*] (H''[\mathbf{\omega}^*])^{-1} \right).
   \label{eq:eig_pixel_final}
\end{equation}
\begin{figure*}[htbp]
    \centering
    \includegraphics[width=1.0\linewidth]{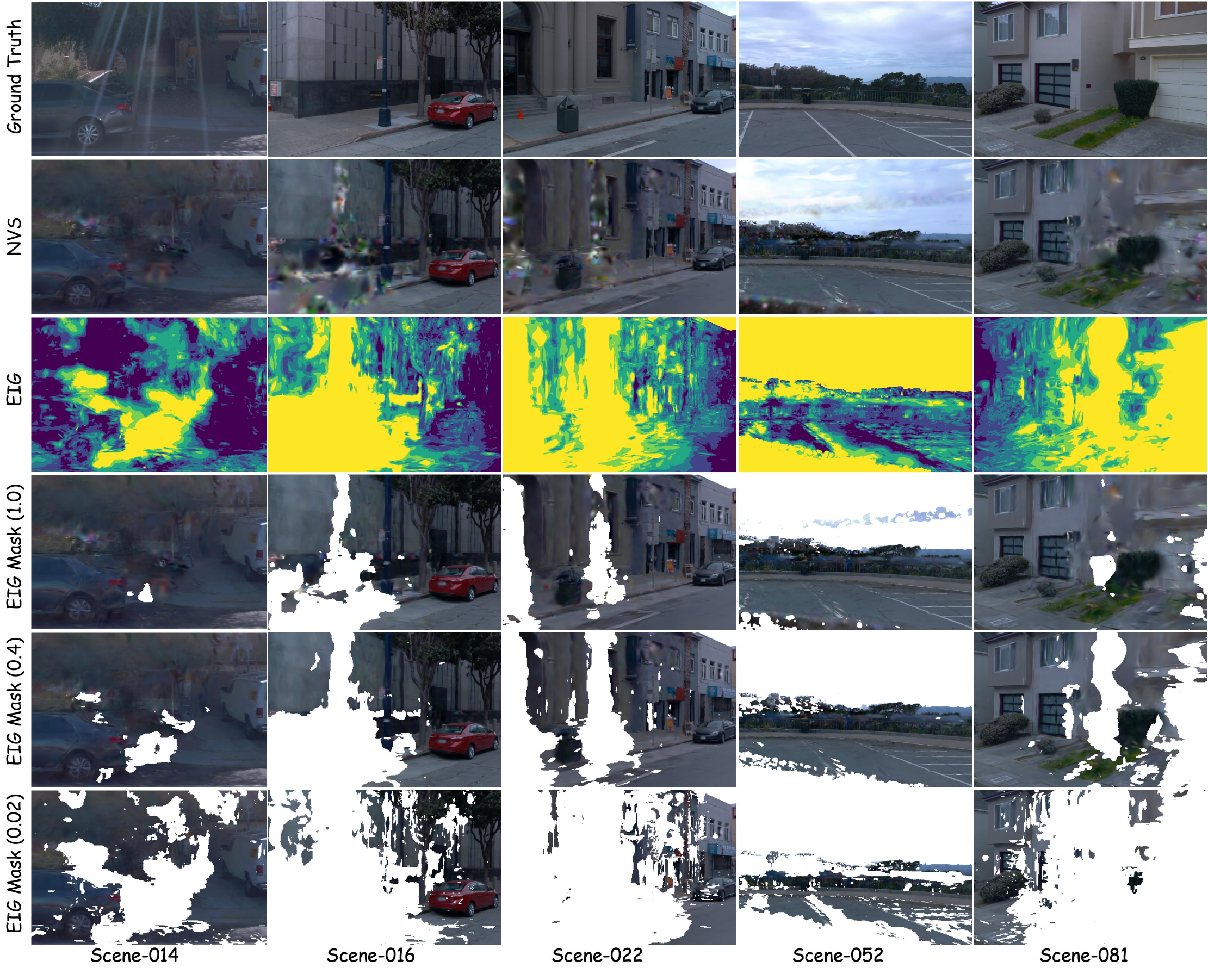}
    \caption{Visualization of EIG as a novel view synthesis quality proxy on representative Waymo \cite{sun2020scalability} scenes. \textbf{Rows (Top to Bottom):} (1) Ground Truth, (2) Novel View Synthesis, (3) Pixel-wise EIG map (yellow = high EIG), (4-6) NVS masked by EIG thresholds ($\tau=1.0, 0.4, 0.02$). White areas are excluded high-EIG pixels, confirming high EIG aligns with NVS artifacts. Sky regions are handled by a separate model and are excluded from this EIG analysis.}
    \label{fig:EIG_mask_supplement}
\end{figure*}

\clearpage
\begin{figure*}[!t]
    \centering
    \includegraphics[width=\linewidth]{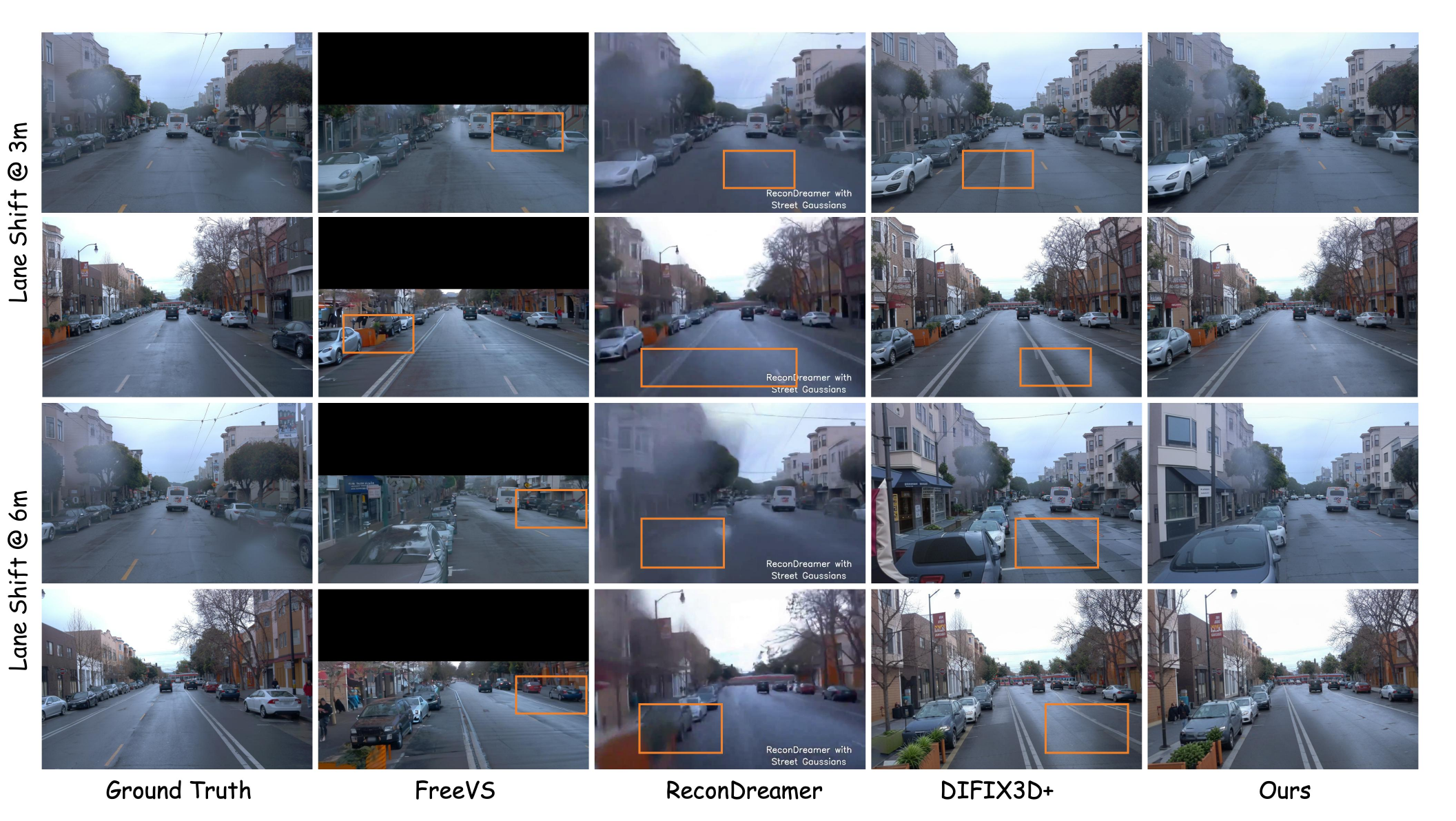}
    \caption{\textbf{Extended Qualitative Comparison on Waymo~\cite{sun2020scalability}.}
    This figure provides additional novel view renderings for the same trajectory across representative methods, complementing the results shown in \cref{fig:metric_gen_contrast} of the main paper. Our method (last column) consistently maintains superior detail and fidelity across challenging regions, highlighted by the orange boxes, compared to methods~\cite{wang2024freevs,ni2025recondreamer,wu2025difix3d+}.}
    \label{fig:metric_gen_contrast_supplement}
\end{figure*}

\section{Additional Visualization Results}
\label{sec:Additional Visualization Results}
\subsection{EIG Correlation Validation and Evaluation Protocol}

As quantified in \cref{fig:EIG_metrics_trend} of the main paper, our cross-camera evaluation validates that pixel-level EIG is highly correlated with NVS quality. We detail the specific evaluation protocol here. 

We first compute co-visible frame sequences between target cameras based on their Field of View (FoV) to ensure sufficient multi-view observation redundancy—a factor critical for optimizing 3D geometry and rendering fidelity. Considering that large low-frequency regions (e.g., solid colors, low-light scenes, or smooth surfaces) disproportionately inflate PSNR scores in standard NVS evaluations, which fails to reflect the model’s ability to capture complex 3D structure, we filtered out frames predominantly containing such low-frequency information when assessing EIG-NVS correlation. This procedure ensures our validation efforts focus on EIG’s efficacy in high-frequency detail and intricate geometry, effectively eliminating the inherent PSNR bias. Following this rigorous filtering process, we identified $4,245$ evaluation pairs for correlation validation.

For qualitative validation, \cref{fig:EIG_mask_supplement} visualizes the EIG map and subsequent masking results on representative Waymo \cite{sun2020scalability} scenes. The figure confirms the correlation: high EIG consistently aligns with NVS artifacts, and progressively masking these high-EIG pixels yields substantially improved perceptual clarity. 
This direct visual evidence not only validates the intuitive link between EIG and synthesis quality but also underscores EIG’s unique advantage as a reliable, pixel-level proxy: it requires no manual annotation of artifacts, operates in a fully unsupervised manner, and provides fine-grained spatial guidance for targeted synthesis refinement.

\subsection{More Qualitative Results}
\begin{figure*}[t]
    \centering
    \includegraphics[width=\linewidth]{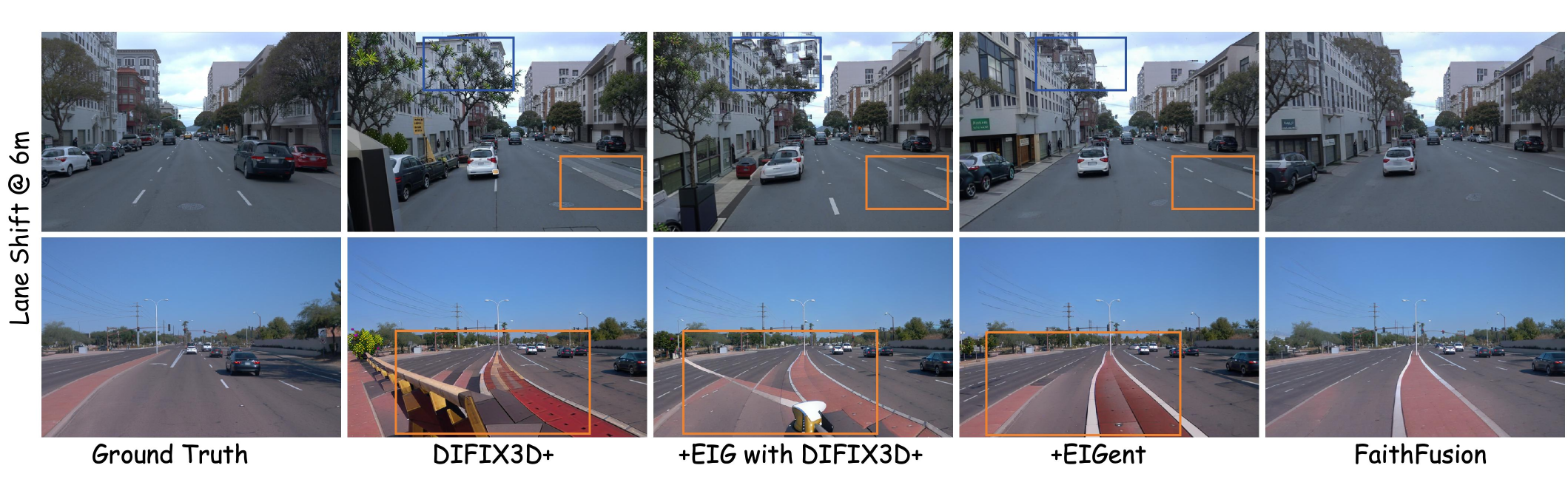}
    \caption{\textbf{Detailed Ablation Study of EIG-Guided Components.}
    This figure provides an extended analysis by quantitatively measuring the incremental performance of integrating $\text{EIG}$-guided components into the $\text{OmniRe}$ baseline (\cref{sec:ablation} in the main paper for the overview). The comprehensive results further confirm the significant role of $\text{EIG}$ guidance in coherently integrating diffusion edits and distilling them back into the 3DGS structure, thus mitigating over-restoration and geometric drift.}
    \label{fig:paper_ablation_supplement}
\end{figure*}

To complement the qualitative analysis in the main paper (\cref{fig:metric_gen_contrast,fig:paper_ablation}), extended visualization results are provided in \cref{fig:metric_gen_contrast_supplement} and 
\cref{fig:paper_ablation_supplement}, where our key conclusions regarding scene synthesis quality and consistency are further validated. All high-resolution visuals and frame-by-frame trajectory comparisons (covering original and novel trajectories with 3 meters/6 meters lane shifts) are available in the accompanying \emph{qualitative\_supplement} folder.

\cref{fig:metric_gen_contrast_supplement} extends the main paper comparisons, showing \emph{FaithFusion} significantly outperforms baselines (~\cite{wang2024freevs,ni2025recondreamer,wu2025difix3d+}) in preserving fine details (e.g., lane markings, building facades) and 3D coherence under large viewpoint shifts. Notably, the black regions in the upper part of the FreeVS~\cite{wang2024freevs} results were manually padded to align with the visualization scale of other methods, as its original output crops the sky region. Our method avoids spurious artifacts (ground bending, semantic mismatches) even at 6 meters lane offsets, aligning with our conclusion that EIG-guided control enables precise "generate-preserve" decisions.

\cref{fig:paper_ablation_supplement} details our ablation results, validating EIG's role as a unified policy that harmonizes diffusion and 3DGS. Consistent with our core insight of replacing heuristics with information-theoretic guidance, EIG suppresses over-restoration in high-confidence regions (preserving 3DGS fidelity) and refines under-constrained areas (enhancing quality). This mechanism resolves the reconstruction-generation trade-off, successfully delivering the three key goals: consistency, quality, and faithfulness.
